\documentclass[10pt,twocolumn,letterpaper]{article}

\usepackage[pagenumbers]{cvpr} % To force page numbers, e.g. for an arXiv version
\usepackage{algorithm}
\usepackage{algpseudocode}
\usepackage{booktabs}   % \toprule, \midrule, \bottomrule, \cmidrule
\usepackage{multirow}   % \multirow
\usepackage{graphicx}   % \rotatebox
\usepackage[table]{xcolor} % \cellcolor and colors

\definecolor{GainFill}{HTML}{D2EAD9}
\definecolor{GainGreen}{HTML}{176B45}
\definecolor{DeltaFill}{HTML}{F0F0F0}
\definecolor{LossRed}{HTML}{B23A48}
\definecolor{LossFill}{HTML}{F6E1E4}

\newcommand{\bestgain}[1]{%
    \cellcolor{GainFill}\textbf{\textcolor{GainGreen}{#1}}}

\newcommand{\deltagain}[1]{%
    \cellcolor{DeltaFill}\textbf{\textcolor{GainGreen}{#1}}}

\newcommand{\deltaloss}[1]{%
    \cellcolor{DeltaFill}\textbf{\textcolor{LossRed}{#1}}}

\newcommand{\deltalabel}[1]{%
    \cellcolor{DeltaFill}\textbf{#1}}

\newcommand{\greenlegend}[1]{%
    \begingroup
    \setlength{\fboxsep}{1.2pt}%
    \colorbox{GainFill}{\textcolor{GainGreen}{\textbf{#1}}}%
    \endgroup}

\newcommand{\graylegend}[1]{%
    \begingroup
    \setlength{\fboxsep}{1.2pt}%
    \colorbox{DeltaFill}{\textbf{#1}}%
    \endgroup}

\newcommand{\redlegend}[1]{%
    \begingroup
    \setlength{\fboxsep}{1.2pt}%
    \colorbox{LossFill}{\textcolor{LossRed}{\textbf{#1}}}%
    \endgroup}

\newcommand{\modelicon}[1]{%
  \raisebox{-0.15\height}{\includegraphics[height=1.15em]{#1}}}

\definecolor{cvprblue}{rgb}{0.21,0.49,0.74}
\usepackage[pagebackref,breaklinks,colorlinks,allcolors=cvprblue]{hyperref}

\def\paperID{*****} % *** Enter the Paper ID here
\def\confName{CVPR}
\def\confYear{2026}

\title{MetaSampling: Making Frame Samplers Efficient for Long-Video Question Answering}

\author{
Ashim Dahal \qquad Bikramjit Banerjee\\
University of Southern Mississippi\\
Hattiesburg, MS, USA\\
{\tt\small \{ashim.dahal,bikramjit.banerjee\}@usm.edu}
}

\begin{document}
% \maketitle
\twocolumn[{%
\renewcommand\twocolumn[1][]{#1}%
\maketitle
\begin{center}
\centering
\captionsetup{type=figure}
\begin{minipage}[t]{0.68\textwidth}
    \centering
    \includegraphics[width=\linewidth]{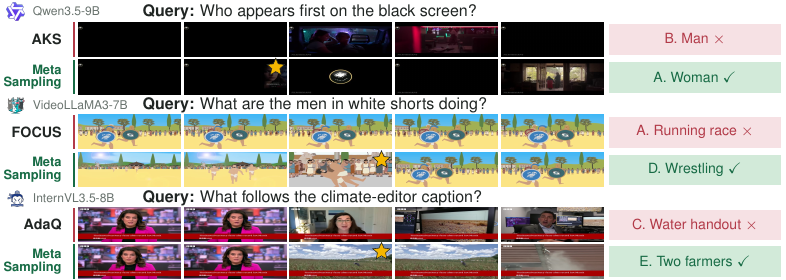}
    \subcaption{LongVideoBench examples across three base selectors and answer models.}
    \label{fig:qual}
\end{minipage}%
\hfill
\begin{minipage}[t]{0.3\textwidth}
    \centering
    \includegraphics[width=\linewidth]{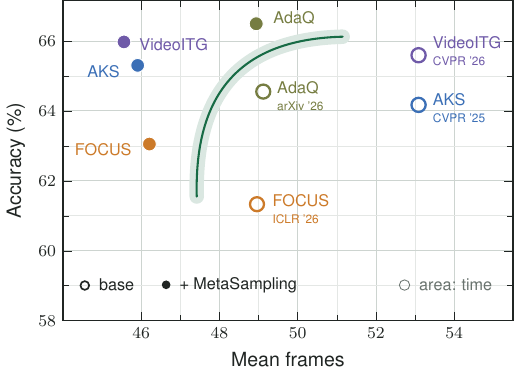}
    \subcaption{LongVideoBench frames vs. accuracy comparison with InternVL3.5-8B.}
    \label{fig:pareto}
\end{minipage}
\caption{MetaSampling improves selected evidence and the accuracy--efficiency trade-off. Gold stars in (a) indicate displayed frames selected only by MetaSampling; marker area in (b) represents mean time per question.}
\label{fig:teaser}
\end{center}%
}]
\begin{abstract}
Frame selection is an important component of long-video question answering (VQA) with Multimodal Large Language Models (MLLMs). Existing frame-selection methods improve over simple top-$k$ embedding retrieval and uniform sampling, but are typically applied under a fixed global selection budget. We introduce \textbf{MetaSampling}, a training-free, plug-and-play sampling strategy that can be applied on top of existing frame selectors. MetaSampling improves downstream VQA efficiency by dynamically reducing the number of frames passed to the MLLM while preserving, and in some cases improving, answer accuracy. We evaluate MetaSampling across 36 paired frame-selector--MLLM-backbone--VQA-benchmark configurations. MetaSampling reduces the number of selected frames in all 36 configurations and improves accuracy in 25 of them, yielding an average frame reduction of $8.9\%$ while slightly improving accuracy overall.

\noindent\emph{Code is available via the supplementary zip.}
\end{abstract}    
\section{Introduction}

%planned paragraph by paragraph
Video Question Answering (VQA) is an important downstream application of Multimodal Large Language Models (MLLMs) or Vision Language Models (VLMs). However, long-context videos incur compute needs beyond a practical budget due to the quadratic complexity of attention. This demands the VQA pipeline to put special emphasis on selection of a smaller subset of frames which carry the most vital information with regard to the user's question.
% long video QA is interesting but attention compute is quadradic so relavent frame selection becomes important

Recent frame-selection methods have achieved strong performance across multiple VQA benchmarks. These approaches span relevance--coverage selection~\cite{tang2025aks,li2026maxinfo,qca2026,fang2026narkfc}, adaptive temporal search~\cite{wang2024videoagent,wang2025videotree,ye2025rethinking,zhu2026focus}, learned instruction-conditioned selection~\cite{hu2025mllm,buch2025flexible,qin2026efficient,wang2026videoitg}, and stochastic or distributional sampling~\cite{liu2025bolt,zhang2026adaq}. Despite their diverse internal representations and selection mechanisms, existing methods typically couple the problem of which frames are informative with the problem of how a finite selection budget is distributed over time: relevance estimation, temporal search, and budget allocation are all determined by the selector's internal procedure. Consequently, improving the allocation of computation or frame budget generally requires modifying the selector itself. This motivates our work in which these two problems are decoupled. Given an arbitrary frozen frame selector, we treat it as a black-box selection primitive and ask how its available budget should be allocated across the video before invoking it locally. This allows for the frame selector's inference behavior to be adapted without changing its scoring function or retraining its parameters.

%In order to do so, w
We abstract the frame selection process %$(Sel_{\psi})$ 
as a function of an array of images $(\mathcal{I})$, question $(q)$ and the selection frame budget constraint $(B)$, $\operatorname{Sel}_{\psi}(\mathcal{I},q;B)$, where $\psi$ is a swappable (base) frame selection method.
% sophisticated solutions like AKS, VITG mostly solve frame selection but come with their own drawbacks.
Building on this abstraction, we propose \textbf{MetaSampling}, a training-free, selector-agnostic strategy that operates on top of an existing frozen selector while leaving its internal scoring or selection mechanism unchanged. MetaSampling uses a small global invocation of the base selector to obtain a coarse query-conditioned estimate of where relevant evidence lies in the video, and uses this estimate to allocate the remaining selection budget across temporal regions before reapplying the same selector locally. In this way, MetaSampling adapts how an existing selector is used and reduces the number of unique frames passed to the downstream MLLM while preserving, and in some cases improving, answer accuracy.

We evaluate MetaSampling on 36 paired MLLM--$Sel_\psi$--benchmark configurations spanning multiple frame-selection methods, MLLMs, and long-video VQA benchmarks. MetaSampling reduces the number of frames passed to the MLLM in 36 of 36 configurations, by $8.9\%$ on average, while improving downstream accuracy in 25 of 36 paired comparisons. A short summary of this Pareto shift between efficiency and accuracy is shown in \cref{fig:pareto}.

To summarize our contributions:

\begin{itemize}
    \item We re-frame sampling as a problem of invoking already sophisticated frozen frame selector $Sel_\psi$, providing a complementary direction to improving $Sel_\psi$ itself.
    \item We introduce MetaSampling: a training-free, selector-agnostic sampling strategy that orchestrates the use of $Sel_\psi$ to improve temporal coverage and allocate dynamic budget while reducing the number of frames passed to the MLLM and preserving accuracy.
    \item Empirically validate the broad generalizability of MetaSampling across selectors, models and benchmarks.
\end{itemize}
% we propose metasampling, a sampling stratedy to adopt on top of these frozen or training free methods that improves efficiency and increases or preservs accuracy for long video qa. It is a simple strategy that can be adopted by a range of selectors and we show consisteng gains and a pareto shift

%our key advantages are
% 1. lower frames 2. preserved accuracy 3. direct result of 1 is lower inference time
\section{Related Work}

\paragraph{Relevance--Coverage Frame Selection.}
A common approach selects frames according to their relevance to the query
while explicitly controlling redundancy or temporal coverage.
AKS~\cite{tang2025aks} formulates keyframe selection by jointly optimizing
query relevance and video coverage, while related methods incorporate
frame diversity, content awareness, or event structure
\cite{fang2026narkfc,li2026maxinfo,qca2026,
zhang2025adardkey,chen2026wfssb,zhang2026cses}.
We use AKS as a representative of this family: it is a strong
training-free, plug-and-play selector with an official public
implementation.

\begin{figure*}[!t]
    \centering
    \includegraphics[width=\linewidth]{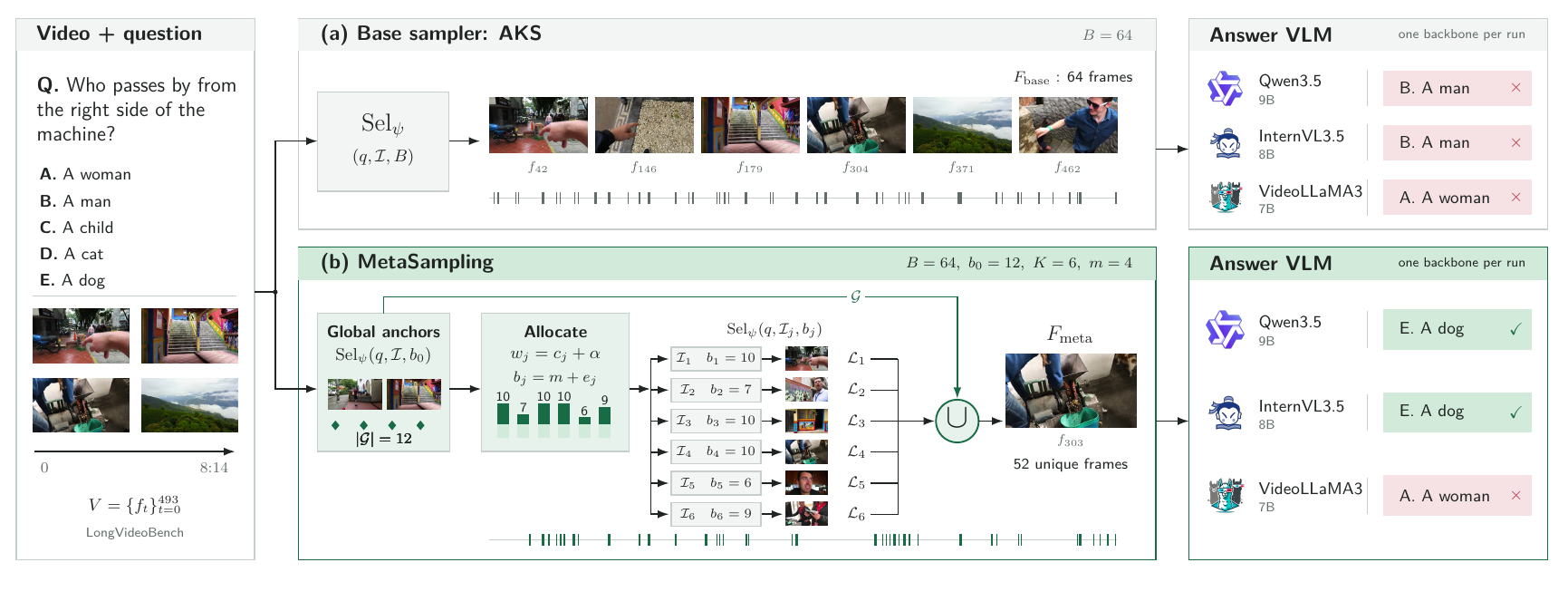}
    \caption{\textbf{Overview of MetaSampling.} Given any frozen frame selector, MetaSampling first obtains a small set of global anchors, uses their temporal distribution to allocate local sampling budgets, applies the same selector independently within each interval, and unions the global and local selections. This preserves broad temporal coverage while concentrating evidence in anchor-relevant regions and can reduce the number of unique frames passed to the answer model.}
    \label{fig:method}
\end{figure*}

\paragraph{Adaptive Temporal Search.}
Rather than scoring a fixed candidate set uniformly, adaptive-search
methods progressively allocate computation to promising temporal regions.
VideoAgent~\cite{wang2024videoagent} performs iterative agentic retrieval,
VideoTree~\cite{wang2025videotree} explores a query-adaptive hierarchical
video representation, and T$^*$~\cite{ye2025rethinking,zhang2026lens}
progressively zooms into relevant temporal regions.
FOCUS~\cite{zhu2026focus} formulates selection as a
bandit-based exploration--exploitation problem over temporal regions.
We adopt FOCUS as a recent training-free representative with an official
public implementation.

\paragraph{Learned Instruction-Conditioned Selection.}
Another line of work learns dedicated selectors from query-conditioned
supervision. M-LLM-based frame selection~\cite{hu2025mllm} uses
MLLM-generated spatial and temporal supervision, while flexible and
reinforcement-learning-based selectors learn policies that optimize
downstream frame utility~\cite{buch2025flexible,qin2026efficient}.
VideoITG~\cite{wang2026videoitg,kim2026request} learns instruction-conditioned temporal
grounding scores for discriminative Top-$K$ frame selection.
We use VideoITG as a representative learned selector; its implementation
and pretrained selector checkpoint are publicly released.

\paragraph{Stochastic and Distributional Sampling.}
Instead of deterministically retaining the highest-scoring frames,
stochastic methods sample according to query-dependent temporal
distributions. BOLT~\cite{liu2025bolt} studies similarity-based sampling
and finds inverse-transform sampling effective for long-video
understanding, while AdaQ~\cite{zhang2026adaq,wang2026evidential} constructs an adaptive
quasi-Gaussian sampling distribution whose concentration varies with the
query. We use AdaQ as the probabilistic representative in our study,
for which a public implementation is available.

\paragraph{Hierarchical and Selector-Refinement Methods.}
Several methods restructure the sampling process rather than relying on a
single flat selection pass. Event-anchored selection constructs
event-level anchors before global refinement~\cite{chen2026efs}, while
hierarchical approaches progressively localize informative temporal
regions~\cite{park2026lvnet,benami2026himu}. Shot-adaptive Frame
Pruning~\cite{zhang2025sophia} post-processes selected keyframes to remove
temporal redundancy. MetaSampling is complementary to these approaches:
it leaves the base selector unchanged and instead performs budget allocation
externally, using a small global selection to assign local interval budgets
before reapplying the same frozen selector. This enables plug-and-play use
across different selector families.

% 02m6ygj2nhk2
\begin{algorithm}[t]
\caption{MetaSampling}
\label{alg:metasampling}
\begin{algorithmic}[1]
\Require $q,\mathcal I,B,B_{\mathrm{new}},b_0,K,m,\alpha,T,T_{\min},
\operatorname{Sel}_{\psi}$
\Ensure $\mathcal F_{\mathrm{out}}$

\If{$T\leq T_{\min}$}
    \State \Return $\operatorname{Sel}_{\psi}(\mathcal I,q;B)$
\EndIf

\State $\mathcal G\gets
       \operatorname{Sel}_{\psi}(\mathcal I,q;b_0)$
\State $(\mathcal I_j)_{j=1}^{K}
       \gets\Call{Partition}{\mathcal I,K}$

\For{$j=1,\ldots,K$}
    \State $w_j\gets
    |\{I_t\in\mathcal G:I_t\in\mathcal I_j\}|+\alpha$
\EndFor

\State $R\gets B_{\mathrm{new}}-b_0$
\State $E\gets R-Km$
\State $(e_1,\ldots,e_K)\gets
       \Call{LargestRemainder}{E,(w_j)_{j=1}^{K}}$
% \State Allocate $E$ across the $K$ intervals proportionally to
%        $(w_j)_{j=1}^{K}$ using largest-remainder rounding
% \State Set $b_j\gets m+e_j$, $j=1,\ldots,K$

\For{$j=1,\ldots,K$}
\State $b_j\gets m+e_j$

    \State $\mathcal L_j\gets
    \operatorname{Sel}_{\psi}(\mathcal I_j,q;b_j)$
\EndFor

\State \Return
$\displaystyle
\mathcal G\cup\bigcup_{j=1}^{K}\mathcal L_j$
\end{algorithmic}
\end{algorithm}
\begin{figure*}[t]
    \centering
    \includegraphics[width=\textwidth]{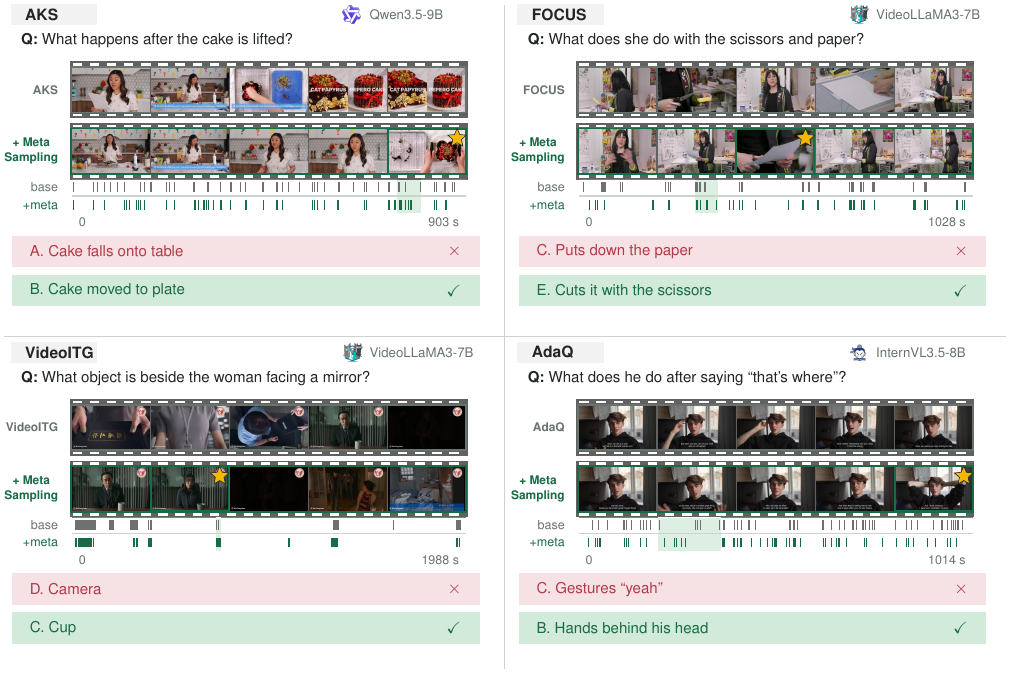}
    \caption{Qualitative LongVideoBench examples for AKS, FOCUS, VideoITG, and AdaQ. Each panel compares frames selected by the base selector (top) and with MetaSampling (bottom). Timelines show the complete selected frame sets and light-green highlights indicate the frames shown in the representative filmstrips. Gold stars mark displayed frames selected only by MetaSampling.}
    \label{fig:qual-detailed}
\end{figure*}

\section{MetaSampling}
\paragraph{Overview.}
Let a video be represented as an array of images
    $\mathcal{I} = \{I_0, I_1, \ldots, I_N\}$.
We denote an arbitrary frozen frame selector by
\begin{equation}
    \mathcal{F}_{\rm{sel}}
    =
    \operatorname{Sel}_{\psi}(\mathcal{I},q;B),
    \qquad
    |\mathcal{F}_{\mathrm{sel}}|\leq B,
    \label{eq:base-selector}
\end{equation}
where $q$ is the question, $B$ is the frame-selection budget, and
$\psi$ denotes the underlying selection method. 

The selected frames $\mathcal{F}_{\mathrm{sel}}$ and the question
$q$ are then provided to an MLLM/VLM for downstream video question answering.
We introduce \textbf{MetaSampling}, a training-free sampling method that
operates on top of an existing frozen selector $\operatorname{Sel}_{\psi}$
under a proposal budget $B_{\mathrm{new}}\leq B$, without modifying its
scoring function or parameters.

\paragraph{Global Relevance Estimation}
We first use a small budget $b_0 \ll B$ to select frames from
the full video. We call these the global selected frames,
\begin{equation}
    \mathcal{G}
    =
    \operatorname{Sel}_{\psi}(\mathcal{I},q;b_0),
    \qquad
    |\mathcal{G}|=b_0.
    \label{eq:global-selection}
\end{equation}

We then partition the video into $K$ uniform, ordered, disjoint temporal intervals
$\{\mathcal I_j\}_{j=1}^{K}$ such that
$\mathcal I=\bigcup_{j=1}^{K}\mathcal I_j$ and
$\mathcal I_j\cap\mathcal I_h=\varnothing$ for $j\neq h$.
For each interval $\mathcal I_j$, we count the global selected frames that
fall within it and use the smoothed count as its relevance weight,
\begin{equation}
\begin{aligned}
    c_j
    &=
    \left|
        \left\{
            I_t\in\mathcal{G}: I_t\in\mathcal{I}_j
        \right\}
    \right|,\\
    w_j
    &=c_j+\alpha,\qquad \alpha>0.
\end{aligned}
\label{eq:interval-relevance}
\end{equation}

\paragraph{Budget Allocation}
After using $b_0$ frames for the global selection, the remaining frame budget is
$R=B_{\mathrm{new}}-b_0$. A minimum budget of $m$ frames is assigned to
each of the $K$ intervals, leaving $E=R-Km$ frames to be redistributed, with
$R\geq Km$.
Interpreting $w_j$ as the relevance of interval $\mathcal I_j$, we formulate
the allocation of the remaining budget as the weighted log-utility problem
% \begin{equation}
% \begin{aligned}
%     \max_{\{u_j\}} \quad&
%     \sum_{j=1}^{K} w_j \log u_j \\
%     \text{s.t.}\quad&
%     \sum_{j=1}^{K} u_j = E,
%     \qquad u_j>0.
% \end{aligned}
% \label{eq:budget-optimization}
% \end{equation}

% \begin{equation}
%     \max_{\{u_j>0\}}
%     \sum_{j=1}^{K} w_j \log u_j
%     \quad
%     \text{s.t.}\quad
%     \sum_{j=1}^{K} u_j = E.
%     \label{eq:budget-optimization}
% \end{equation}

\begin{equation}
    \max_{\{u_j\}}
    \sum_{j=1}^{K} w_j \log u_j
    \quad
    \text{s.t.}\quad
    \sum_{j=1}^{K} u_j = E,\;\; u_j>0.
    \label{eq:budget-optimization}
\end{equation}
The continuous solution allocates the $E$ remaining frames proportionally
to the interval relevance weights,
\begin{equation}
    u_j
    =
    E\frac{w_j}{\sum_{h=1}^{K}w_h}.
    \label{eq:proportional-budget}
\end{equation}
The fractional allocations $\{u_j\}$ are converted to integer frame counts
$(e_j)$ using the largest-remainder method, and the final interval budget is
$b_j=m+e_j$, such that $\sum_{j=1}^{K}b_j=R$.

\paragraph{Local Selection}
For each temporal interval $\mathcal I_j$, we apply the same selector using
its allocated budget $b_j$,
\begin{equation}
    \mathcal L_j
    =
    \operatorname{Sel}_{\psi}(\mathcal I_j,q;b_j),
    \qquad j=1,\ldots,K.
    \label{eq:interval-selection}
\end{equation}
The interval selections are combined with the global selected frames as
\begin{equation}
    \mathcal F_{\mathrm{meta}}
    =
    \mathcal G
    \cup
    \bigcup_{j=1}^{K}\mathcal L_j,
    \qquad
    |\mathcal F_{\mathrm{meta}}|
    \leq B_{\mathrm{new}}\leq B.
    \label{eq:metasampling-union}
\end{equation}

MetaSampling is applied only to videos longer than a duration threshold
$T_{\min}$. For videos with $T\leq T_{\min}$, we use the base selector
$\operatorname{Sel}_{\psi}(\mathcal I,q;B)$ directly; otherwise, we use
$\mathcal F_{\mathrm{meta}}$.

The complete procedure is summarized in \cref{alg:metasampling}, with the
visual workflow shown in \cref{fig:method}.
\section{Experiments}
\begin{table*}[!t]
\centering
\small
\setlength{\tabcolsep}{1.2pt}
\renewcommand{\arraystretch}{1.18}

\caption{Accuracy (\%), inference time (s/q), and mean number of frames
passed to the answer model for four selectors across three MLLM/VLM
backbones. Overall columns are weighted by the number of evaluated questions
in each dataset (2,700 VideoMME, 1,337 LongVideoBench, and 1,549 LVBench).
The $\Delta$ Avg. row reports mean accuracy change, time speedup,
and relative frame-count change. \greenlegend{Green} cells mark paired
improvements of at least $1.0$ percentage point in accuracy or $1.15\times$
in speed; \redlegend{red} cells mark paired accuracy losses greater than
$1.0$ percentage point; \graylegend{gray} cells denote per-column aggregate
statistics per backbone. $\dagger$ denotes a probabilistic selector.}
\label{tab:main-results}

\begin{tabular}{@{}cl*{12}{c}@{}}
\toprule
\multirow{2}{*}{Model}
& \multirow{2}{*}{Method}
& \multicolumn{3}{c}{VideoMME~\cite{fu2025videomme}}
& \multicolumn{3}{c}{LongVideoBench~\cite{wu2024longvideobench}}
& \multicolumn{3}{c}{LVBench~\cite{wang2025lvbench}}
& \multicolumn{3}{c}{Overall} \\
\cmidrule(lr){3-5}
\cmidrule(lr){6-8}
\cmidrule(lr){9-11}
\cmidrule(lr){12-14}
& & Acc. $\uparrow$ & Time $\downarrow$ & \# Frames $\downarrow$
& Acc. $\uparrow$ & Time $\downarrow$ & \# Frames $\downarrow$
& Acc. $\uparrow$ & Time $\downarrow$ & \# Frames $\downarrow$
& Acc. $\uparrow$ & Time $\downarrow$ & \# Frames $\downarrow$ \\
\midrule

% ====================== Qwen3.5-9B ======================
\multirow[c]{9}{*}{%
  \rotatebox[origin=c]{90}{%
    \modelicon{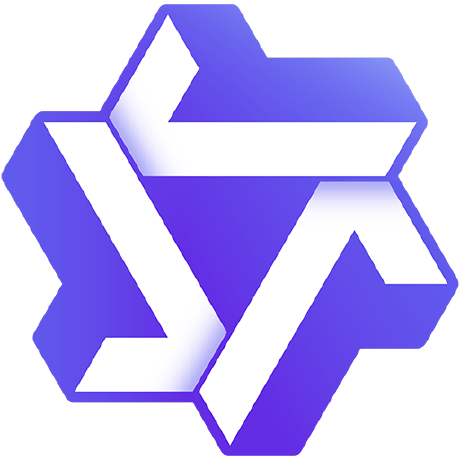}\,
    Qwen3.5-9B~\cite{qwen2026qwen35}}}

& AKS~\cite{tang2025aks}
& 69.81 & 7.92 & 62.28
& 67.46 & 11.71 & 53.09
& 51.52 & 13.10 & 64.00
& 64.18 & 10.26 & 60.56 \\

& AKS + ours
& 69.96
& \bestgain{6.82}
& 54.91
& \bestgain{68.51}
& 10.59
& 45.91
& 51.97
& \bestgain{10.47}
& 52.12
& 64.62
& \bestgain{8.73}
& 51.98 \\

\cmidrule(l){2-14}

& FOCUS~\cite{zhu2026focus}
& 66.81 & 7.13 & 55.52
& 64.85 & 11.16 & 48.96
& 52.55 & 13.10 & 64.00
& 62.39 & 9.75 & 56.30 \\

& FOCUS + ours
& 67.37 & 6.73 & 53.15
& 65.07 & 10.75 & 46.21
& 53.07 & 12.42 & 61.08
& 62.85 & 9.27 & 53.69 \\

\cmidrule(l){2-14}

& VideoITG~\cite{wang2026videoitg}
& 73.04 & 7.94 & 62.28
& 70.91 & 11.68 & 53.09
& 59.07 & 13.08 & 64.00
& 68.66 & 10.26 & 60.56 \\

& VideoITG + ours
& 72.59
& \bestgain{6.86}
& 54.77
& 70.31
& 10.48
& 45.56
& \cellcolor{LossFill}\textbf{\textcolor{LossRed}{57.33}}
& \bestgain{10.54}
& 52.00
& 67.81
& \bestgain{8.75}
& 51.80 \\

\cmidrule(l){2-14}

& AdaQ$^\dagger$~\cite{zhang2026adaq}
& 69.37 & 7.52 & 59.63
& 70.38 & 11.12 & 49.12
& 52.74 & 12.90 & 62.90
& 65.00 & 9.87 & 58.02 \\

& AdaQ$^\dagger$ + ours
& 70.30 & 7.48 & 58.76
& \cellcolor{LossFill}\textbf{\textcolor{LossRed}{68.44}}
& 11.10
& 48.94
& 53.13 & 12.66 & 61.73
& 65.09 & 9.78 & 57.23 \\

\cmidrule(l){2-14}

& \deltalabel{$\Delta$ Avg.}
& \deltagain{+0.30}
& \deltagain{$1.10\times$}
& \deltagain{-7.4\%}
& \deltaloss{-0.32}
& \deltagain{$1.07\times$}
& \deltagain{-8.4\%}
& \deltaloss{-0.10}
& \deltagain{$1.14\times$}
& \deltagain{-10.9\%}
& \deltagain{+0.04}
& \deltagain{$1.10\times$}
& \deltagain{-8.8\%} \\

\midrule

% ====================== InternVL3.5-8B ======================
\multirow[c]{9}{*}{%
  \rotatebox[origin=c]{90}{%
    \modelicon{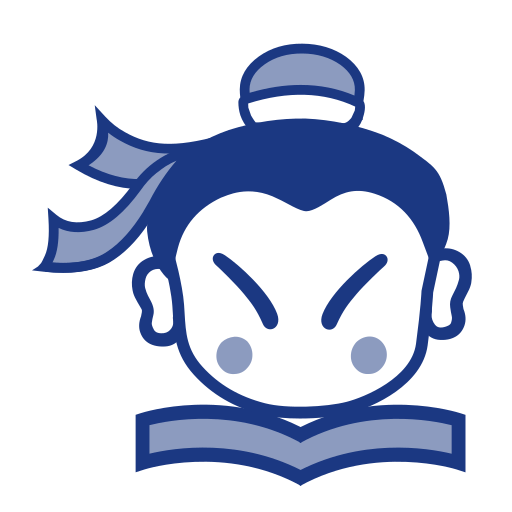}\,
    InternVL3.5-8B~\cite{wang2025internvl35}}}

& AKS
& 68.56 & 2.91 & 62.28
& 64.17 & 6.89 & 53.09
& 47.39 & 6.03 & 64.00
& 61.64 & 4.73 & 60.56 \\

& AKS + ours
& 68.96
& \bestgain{2.48}
& 54.91
& \bestgain{65.30}
& 6.54
& 45.91
& \bestgain{50.74}
& \bestgain{4.88}
& 52.12
& \bestgain{63.03}
& 4.12
& 51.98 \\

\cmidrule(l){2-14}

& FOCUS
& 66.48 & 2.67 & 55.52
& 61.33 & 6.69 & 48.96
& 49.84 & 6.07 & 64.00
& 60.63 & 4.58 & 56.30 \\

& FOCUS + ours
& 67.37
& 2.48
& 53.15
& \bestgain{63.05}
& 6.56
& 46.21
& 49.26
& 5.86
& 61.08
& 61.31 & 4.39 & 53.69 \\

\cmidrule(l){2-14}

& VideoITG
& 72.00 & 2.86 & 62.28
& 65.59 & 6.87 & 53.09
& 54.10 & 5.97 & 64.00
& 65.50 & 4.68 & 60.56 \\

& VideoITG + ours
& 71.26
& \bestgain{2.42}
& 54.77
& 65.97
& 6.47
& 45.56
& 53.58
& \bestgain{4.74}
& 52.00
& 65.09
& \bestgain{4.03}
& 51.80 \\

\cmidrule(l){2-14}

& AdaQ$^\dagger$
& 69.07 & 2.82 & 59.63
& 64.55 & 6.68 & 49.12
& 48.87 & 5.99 & 62.90
& 62.39 & 4.62 & 58.02 \\

& AdaQ$^\dagger$ + ours
& \bestgain{70.26}
& 2.77
& 58.76
& \bestgain{66.49}
& 6.70
& 48.94
& \bestgain{49.90}
& 5.91
& 61.73
& \bestgain{63.71}
& 4.58
& 57.23 \\

\cmidrule(l){2-14}

& \deltalabel{$\Delta$ Avg.}
& \deltagain{+0.44}
& \deltagain{$1.11\times$}
& \deltagain{-7.4\%}
& \deltagain{+1.29}
& \deltagain{$1.03\times$}
& \deltagain{-8.4\%}
& \deltagain{+0.82}
& \deltagain{$1.14\times$}
& \deltagain{-10.9\%}
& \deltagain{+0.75}
& \deltagain{$1.09\times$}
& \deltagain{-8.8\%} \\

\midrule

% ====================== VideoLLaMA3-7B ======================
\multirow[c]{9}{*}{%
  \rotatebox[origin=c]{90}{%
    \modelicon{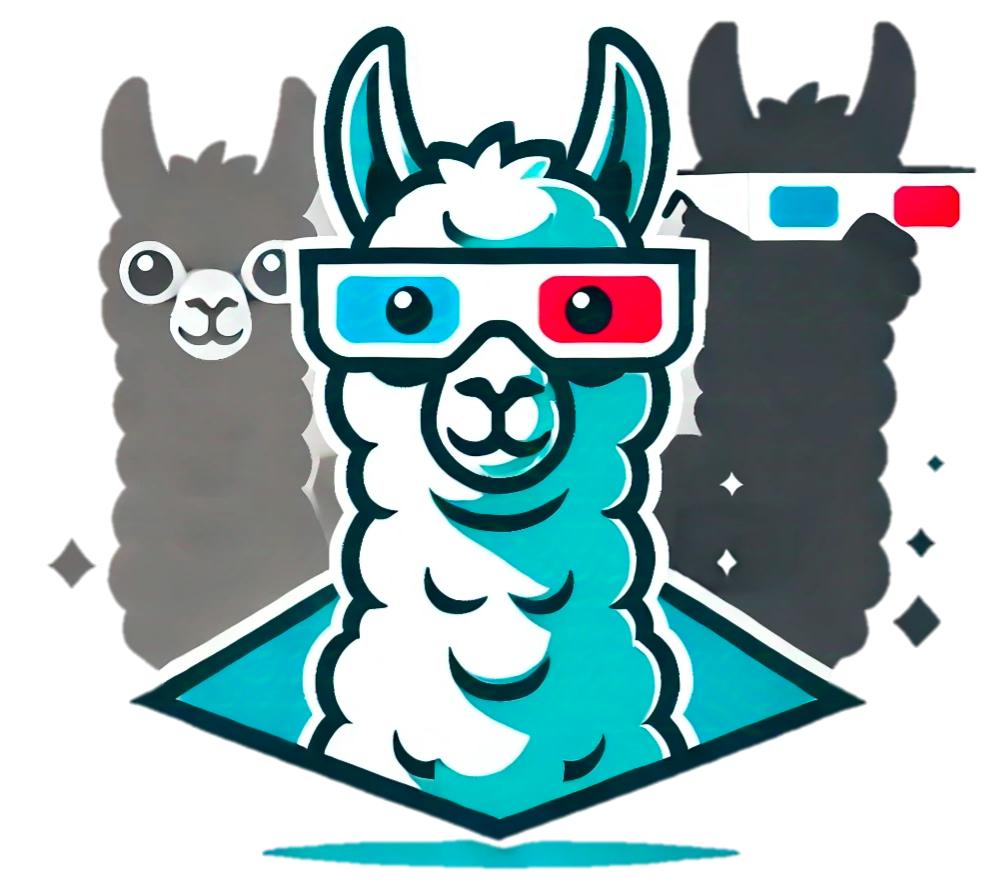}\,
    VideoLLaMA3-7B~\cite{zhang2025videollama3}}}

& AKS
& 57.44 & 2.80 & 62.28
& 48.69 & 6.89 & 53.09
& 41.32 & 5.68 & 64.00
& 50.88 & 4.58 & 60.56 \\

& AKS + ours
& 57.52 & 2.68 & 54.91
& 47.94 & 6.81 & 45.91
& \bestgain{43.32}
& 5.09 & 52.12
& 51.29 & 4.34 & 51.98 \\

\cmidrule(l){2-14}

& FOCUS
& 55.63 & 2.91 & 55.52
& 48.17 & 6.62 & 48.96
& 44.16 & 5.78 & 64.00
& 50.66 & 4.59 & 56.30 \\

& FOCUS + ours
& \bestgain{57.41}
& 2.66
& 53.15
& \cellcolor{LossFill}\textbf{\textcolor{LossRed}{46.52}}
& 6.57
& 46.21
& 44.22
& 5.50
& 61.08
& 51.15 & 4.38 & 53.69 \\

\cmidrule(l){2-14}

& VideoITG
& 61.26 & 2.80 & 62.28
& 48.77 & 6.91 & 53.09
& 46.42 & 5.69 & 64.00
& 54.16 & 4.59 & 60.56 \\

& VideoITG + ours
& \cellcolor{LossFill}\textbf{\textcolor{LossRed}{59.85}}
& 2.67
& 54.77
& 49.36
& 6.81
& 45.56
& 47.19
& 5.09
& 52.00
& 53.83 & 4.33 & 51.80 \\

\cmidrule(l){2-14}

& AdaQ$^\dagger$
& 57.48 & 2.75 & 59.63
& 48.62 & 6.85 & 49.12
& 43.12 & 5.85 & 62.90
& 51.38 & 4.59 & 58.02 \\

& AdaQ$^\dagger$ + ours
& 58.37
& 2.71
& 58.76
& 48.09
& 6.83
& 48.94
& 43.64
& 5.75
& 61.73
& 51.82 & 4.54 & 57.23 \\

\cmidrule(l){2-14}

& \deltalabel{$\Delta$ Avg.}
& \deltagain{+0.33}
& \deltagain{$1.05\times$}
& \deltagain{-7.4\%}
& \deltaloss{-0.58}
& \deltagain{$1.01\times$}
& \deltagain{-8.4\%}
& \deltagain{+0.84}
& \deltagain{$1.08\times$}
& \deltagain{-10.9\%}
& \deltagain{+0.25}
& \deltagain{$1.04\times$}
& \deltagain{-8.8\%} \\

\bottomrule
\end{tabular}
\end{table*}
\begin{figure*}[t]
    \centering
    \includegraphics[width=\textwidth]{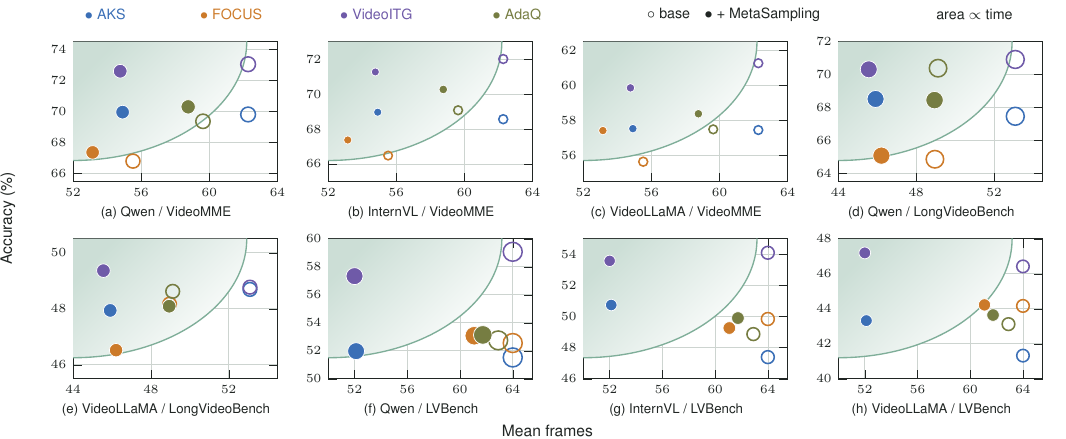}
    \caption{Accuracy--efficiency trade-offs across eight model--dataset settings. Ninth one is on \cref{fig:pareto}. The green shading indicates the favorable upper-left direction. Axis scales vary by panel.}
    \label{fig:pareto-eight}
\end{figure*}
\subsection{Configuration}
\label{sec:expdetail}
\paragraph{MLLM/VLM Backbones.}
We evaluate MetaSampling with three multimodal backbones:
Qwen3.5-9B~\cite{qwen2026qwen35},
InternVL3.5-8B~\cite{wang2025internvl35}, and
VideoLLaMA3-7B~\cite{zhang2025videollama3}.
Qwen and Intern backbones were hosted with VLLM and VideoLLaMA was hosted within the Transformers library.

\paragraph{Frame Selection Methods}
We evaluate MetaSampling with four diverse frame selection methods:
Adaptive Keyframe Sampling (AKS)~\cite{tang2025aks},
FOCUS~\cite{zhu2026focus},
VideoITG~\cite{wang2026videoitg}, and
AdaQ~\cite{zhang2026adaq}.
AKS, FOCUS, and VideoITG are evaluated deterministically using fixed precomputed embeddings or selection scores and zero temperature, while AdaQ samples frames probabilistically from its sampling distribution.

\paragraph{Datasets}
We evaluate on three long-video question-answering benchmarks:
Video-MME~\cite{fu2025videomme},
LongVideoBench~\cite{wu2024longvideobench}, and
LVBench~\cite{wang2025lvbench}.
Across the evaluated splits, the three benchmarks comprise 5,586 question--answer pairs: 2,700 from Video-MME, 1,337 from LongVideoBench, and 1,549 from LVBench. For LongVideoBench, we evaluate on its official validation split of 1,337 questions over 752 videos.

\paragraph{Hyperparameters}
Unless otherwise stated, we use $B_{\mathrm{new}} = B=64$, $b_0=12$, $K=6$, $m=4$, and $\alpha=1$. MetaSampling is applied to videos longer than $T_{\min}=300$ seconds (5 minutes), and videos are decoded at one frame per second for all experiments.

\paragraph{Evaluation Protocol}
For efficiency, we report the mean number of frames passed to the MLLM and
the mean inference time per question. For Qwen3.5-9B, timing includes frame
selection, evidence loading, and answer-model inference. For InternVL3.5-8B
and VideoLLaMA3-7B, cached selector outputs are reused, and timing therefore
includes only answer-model inference. Timing values are therefore interpreted only within each backbone.

\subsection{Results}

\paragraph{Across Selectors and Backbones}
We present the primary results of 72 evaluations (3 datasets, 3 MLLM
backbones, 4 selector methods, and 2 variants each), forming 36 matched
base--MetaSampling comparisons, in \cref{tab:main-results}. Overall,
MetaSampling reduces the number of frames passed to the answer
model by 8.9\% on average while improving accuracy in 25/36 comparisons,
with gains of up to +3.35 percentage points. Question-weighted accuracy changes are $+0.04$, $+0.75$, and $+0.25$ points across Qwen3.5-9B, InternVL3.5-8B, and VideoLLaMA3-7B. Inference time decreases
in 35/36 comparisons under the backbone-specific timing protocols described
in \cref{sec:expdetail}. Among the 11 accuracy decreases, only 4 exceed
$1.0$ percentage point; frame count is reduced in all 36 comparisons.

An interesting observation of \cref{tab:main-results} is that frame selection is not information-monotonic: retaining more apparently relevant frames does not necessarily produce a better answer. For example in AKS with LVBench and InternVL3.5-8B backbone, MetaSampling increases accuracy by $+3.35$-points while using $18.6\%$ fewer frames. This suggests that efficiency and answer quality need not be competing objectives. An illustrative comparison of this result is also shown in \cref{fig:pareto-eight}.

\paragraph{Matched-Budget Results}

\begin{table}[t]
\centering
\setlength{\tabcolsep}{4pt}
\renewcommand{\arraystretch}{1.2}
\caption{Matched-budget accuracy (\%) with Qwen3.5-9B. Budgets are matched by
mean evidence count from cached selector outputs; AdaQ retains its 64-proposal
runs. Gray $\Delta$ rows report paired accuracy changes, with green and red
denoting gains and losses, respectively. $\dagger$ denotes a probabilistic
sampler.}
\label{tab:matchedbudget}

\begin{tabular}{@{}lccc@{}}

\toprule
Method & VMME & LVB & LVBench \\
\midrule

AKS
& 69.67
& 67.61
& 50.23 \\

AKS + ours
& 69.96
& 68.51
& 51.97 \\

\deltalabel{$\Delta$}
& \deltagain{+0.29}
& \deltagain{+0.90}
& \deltagain{+1.74} \\

\midrule

FOCUS
& 66.81
& 65.82
& 52.36 \\

FOCUS + ours
& 67.37
& 65.07
& 53.07 \\

\deltalabel{$\Delta$}
& \deltagain{+0.56}
& \deltaloss{-0.75}
& \deltagain{+0.71} \\

\midrule

VideoITG
& 72.59
& 70.31
& 56.94 \\

VideoITG + ours
& 72.59
& 70.31
& 57.33 \\

\deltalabel{$\Delta$}
& \deltalabel{0.00}
& \deltalabel{0.00}
& \deltagain{+0.39} \\

\midrule

AdaQ$^\dagger$
& 69.37
& 70.38
& 52.74 \\

AdaQ + ours
& 70.30
& 68.44
& 53.13 \\

\deltalabel{$\Delta$}
& \deltagain{+0.93}
& \deltaloss{-1.94}
& \deltagain{+0.39} \\

\bottomrule
\end{tabular}
\end{table}

We isolate MetaSampling's accuracy gains from its reduction in evidence count through the matched-budget comparison in \cref{tab:matchedbudget}. The resulting proposal budgets are $(56,54,52)$ for AKS, $(60,60,61)$ for FOCUS, and $(56,54,52)$ for VideoITG on (VideoMME, LongVideoBench, LVBench), respectively; AdaQ already matches within one frame.

Under matched budgets, MetaSampling improves accuracy in 8/12 comparisons, ties in 2, and decreases in only 2, for an average gain of $+0.27$ points. The advantage is strongest for AKS, with gains of $+0.29$, $+0.90$, and $+1.74$ points across the three datasets. These results show that MetaSampling improves \emph{which} evidence is retained, not merely how much.

\paragraph{Qualitative Results}
Qualitative examples across the four selectors and three MLLM backbones
are presented in \cref{fig:qual,fig:qual-detailed}.
\subsection{Ablation Study}

\paragraph{Backfill Frames}

% ====================== E2: backfill control ======================
\begin{table}[t]
\centering
\small
\setlength{\tabcolsep}{3.2pt}
\renewcommand{\arraystretch}{1.15}
\caption{Backfill control on LongVideoBench using Qwen3.5-9B and AKS.
Route metrics use the same 839 MetaSampling-route questions; full accuracy
includes the 498 unchanged short-route questions. \greenlegend{Green} cells
mark the better value for each metric; the \graylegend{$\Delta$} row reports
backfill relative to no backfill.}
\label{tab:metasampling-backfill-ablation}

\begin{tabular}{@{}lcccc@{}}
\toprule
Method
& \shortstack{Route\\Acc. (\%)}
& \shortstack{Full\\Acc. (\%)}
& \shortstack{Time\\(s/q) $\downarrow$}
& \shortstack{Frames\\$\downarrow$} \\
\midrule

Plain AKS
& 63.05
& 67.46
& 15.19
& 64.00 \\

Ours, no backfill
& 64.72
& 68.51
& \bestgain{13.40}
& \bestgain{52.55} \\

Ours, backfill
& \bestgain{65.55}
& \bestgain{69.04}
& 15.20
& 64.00 \\

\midrule

\deltalabel{$\Delta$}
& \deltagain{$+0.83$}
& \deltagain{$+0.52$}
& \deltaloss{$+13.4\%$}
& \deltaloss{$+21.8\%$} \\

\bottomrule
\end{tabular}
\end{table}
Similar to the matched-budget setting, we test the effect of utilizing the entire budget $(B)$ for each question. To test this, we remove the $\mathcal{G}$ globally selected frames from the array of images $\mathcal{I}$. This causes the union operation in \cref{eq:metasampling-union} to have no effect on the final selected frames $\mathcal F_{\mathrm{out}}$. We refer to these additional frames as backfilled frames and report the results in \cref{tab:metasampling-backfill-ablation}.

Backfilling the remaining budget improves route and full accuracy by \(0.83\) and \(0.52\) points, respectively, but increases inference time by \(13.4\%\) and frame usage by \(21.8\%\) compared to no backfill. MetaSampling thus improves over the plain AKS selector both with and without backfilling for maximum budget utilization.

\paragraph{Anchor Weights}
We study the effect of the anchor relevance weighting in \cref{eq:interval-relevance} on the interval allocation in \cref{eq:interval-selection}, particularly the choice of per-interval budget $(b_j)$. As shown in \cref{tab:metasampling-allocation-ablation}, anchor-weighted allocation improves route accuracy by $+2.26$ points and full accuracy by $+1.42$ points over equal allocation, while also slightly reducing inference time and frame count. This confirms that allocating interval budgets according to anchor relevance is more effective than distributing them uniformly.

% ====================== E1: allocation ablation ======================
\begin{table}[t]
\centering
\small
\setlength{\tabcolsep}{3.5pt}
\renewcommand{\arraystretch}{1.15}
\caption{Allocation ablation on LongVideoBench using Qwen3.5-9B and AKS.
Route accuracy is measured on the 839 MetaSampling-route questions; full
accuracy includes the 498 unchanged short-route questions. Both variants
disable backfill. \greenlegend{Green} cells mark the better paired value;
the \graylegend{$\Delta$} row reports anchor-weighted minus equal allocation.}
\label{tab:metasampling-allocation-ablation}

\begin{tabular}{@{}lcccc@{}}
\toprule
Allocation
& \shortstack{Route\\Acc. (\%)}
& \shortstack{Full\\Acc. (\%)}
& \shortstack{Time\\(s/q) $\downarrow$}
& \shortstack{Frames\\$\downarrow$} \\
\midrule

Equal
& 62.46
& 67.09
& 13.48
& 52.91 \\

Anchor-weighted (ours)
& \bestgain{64.72}
& \bestgain{68.51}
& \bestgain{13.40}
& \bestgain{52.55} \\

\midrule

\deltalabel{$\Delta$}
& \deltagain{$+2.26$}
& \deltagain{$+1.42$}
& \deltagain{$1.01\times$}
& \deltagain{$-0.68\%$} \\

\bottomrule
\end{tabular}
\end{table}

\subsection{Analysis and Discussion}
\subsubsection{Hyperparameter Sensitivity}

% ====================== E3: full-benchmark b0 sensitivity ======================
\begin{table}[t]
\centering
\footnotesize
\setlength{\tabcolsep}{3.2pt}
\renewcommand{\arraystretch}{1.15}
\caption{Full-LongVideoBench sensitivity to the global-anchor budget $b_0$
using Qwen3.5-9B and AKS. Route metrics use all 839 union-route questions;
full accuracy includes the 498 unchanged short-route questions. All other
settings are fixed at $B=64$, $K=6$, $m=4$, and $\alpha=1$. Confidence
intervals use 10,000 video-level bootstrap resamples over 326 videos. Paired
differences and W/L/T are reported as the default $b_0=12$ minus each
alternative.}
\label{tab:giu-global-budget-sensitivity}
\begin{tabular}{@{}ccccc@{}}
\toprule
$b_0$
& \shortstack{Route\\Acc. (\%)}
& \shortstack{Full\\Acc. (\%)}
& \shortstack{Time\\(s/q) $\downarrow$}
& \shortstack{Frames\\$\downarrow$} \\
\midrule
4
& \bestgain{64.84}
& \bestgain{68.59}
& 14.69
& 60.09 \\
\textbf{12 }
& 64.72
& 68.51
& 13.40
& 52.55 \\
20
& 61.86
& 66.72
& \bestgain{12.53}
& \bestgain{46.98} \\
\bottomrule
\end{tabular}

\vspace{4pt}

\setlength{\tabcolsep}{2.0pt}
\begin{tabular}{@{}lccc@{}}
\toprule
Comparison
& \shortstack{$\Delta$ Route\\Acc. (pp)}
& 95\% CI
& W/L/T \\
\midrule
$b_0=12$ vs. $b_0=4$
& \deltaloss{$-0.12$}
& $[-1.89,+1.64]$
& 29/30/780 \\
$b_0=12$ vs. $b_0=20$
& \deltagain{$+2.86$}
& $[+1.18,+4.63]$
& 42/18/779 \\
\bottomrule
\end{tabular}
\end{table}

\paragraph{Global Frame Budget ($b_0$) Sensitivity}
Varying the global-frame budget shows that MetaSampling is generally robust to moderate changes in $b_0$. In particular, $b_0=4$ and $b_0=12$ achieve nearly identical route accuracy, with a paired difference of only $-0.12$ points and a confidence interval spanning zero. The W/L/T counts (win/loss/tie) are also nearly balanced at $29/30/780$, indicating that most questions are unaffected. Increasing $b_0$ to $20$ reduces route accuracy by $2.86$ points relative to $b_0=12$, while lowering inference time and frame usage.

\paragraph{Video Intervals ($K$) Sensitivity}
MetaSampling is similarly stable across the tested numbers of temporal intervals. Relative to $K=6$, using $K=4$ or $K=8$ changes route accuracy by only $1.79$ and $1.19$ points, respectively, with both confidence intervals spanning zero. The corresponding W/L/T counts are dominated by ties ($758$ and $767$ of $839$ questions), while inference time and frame usage remain nearly unchanged, indicating limited sensitivity to $K$ over this range.

% ====================== K sensitivity ======================
\begin{table}[t]
\centering
\footnotesize
\setlength{\tabcolsep}{3.2pt}
\renewcommand{\arraystretch}{1.15}
\caption{Full-LongVideoBench sensitivity to the number of temporal intervals
$K$ using Qwen3.5-9B and AKS. Route metrics use all 839 union-route questions;
full accuracy includes the 498 unchanged short-route questions. All settings
use $B=64$, $b_0=12$, $m=4$, and $\alpha=1$. Confidence intervals use 10,000
video-level bootstrap resamples over 326 videos. Paired differences and W/L/T
are reported as the default $K=6$ minus each alternative.}
\label{tab:giu-k-sensitivity}
\begin{tabular}{@{}ccccc@{}}
\toprule
$K$
& \shortstack{Route\\Acc. (\%)}
& \shortstack{Full\\Acc. (\%)}
& \shortstack{Time\\(s/q) $\downarrow$}
& \shortstack{Frames\\$\downarrow$} \\
\midrule
4
& 62.93
& 67.39
& \bestgain{13.33}
& \bestgain{52.41} \\
\textbf{6 }
& \bestgain{64.72}
& \bestgain{68.51}
& 13.40
& 52.55 \\
8
& 63.53
& 67.76
& 13.45
& 52.84 \\
\bottomrule
\end{tabular}

\vspace{4pt}

\setlength{\tabcolsep}{2.0pt}
\begin{tabular}{@{}lccc@{}}
\toprule
Comparison
& \shortstack{$\Delta$ Route\\Acc. (pp)}
& 95\% CI
& W/L/T \\
\midrule
$K=6$ vs. $K=4$
& \deltagain{$+1.79$}
& $[-0.35,+3.95]$
& 48/33/758 \\
$K=6$ vs. $K=8$
& \deltagain{$+1.19$}
& $[-0.83,+3.25]$
& 41/31/767 \\
\bottomrule
\end{tabular}
\end{table}

\subsubsection{Recommended Hyperparameter Selection.}
The three hyperparameters $m$, $K$, and $b_0$ control complementary
aspects of MetaSampling. We recommend first choosing $K$ according to
video length. Smaller $K$ produces coarser temporal regions with larger
per-interval budgets, while larger $K$ provides finer temporal
localization at the cost of dividing the available budget across more
regions. In practice, we find a modest range of
$K\in[2^2,2^3]=[4,8]$ sufficient, using fewer intervals for shorter
videos and more intervals as video duration increases. This is also
well matched to selectors such as AKS~\cite{tang2025aks}, which already
perform recursive temporal refinement within each interval. Our
sensitivity results show limited variation across $K\in\{4,6,8\}$,
with $K=6$ used as the default.

Given $K$, we set the minimum interval budget $m$ such that approximately
$40$--$50\%$ of the remaining budget $R=B_{\mathrm{new}}-b_0$ is
reserved for uniform temporal coverage,
\[
    Km \approx (0.4\text{--}0.5)R,
\]
leaving the remaining frames to be allocated according to anchor
relevance. Our default $m=4$ with $K=6$ reserves $24/52=46.2\%$ of
$R$ for this coverage floor. Finally, $b_0$ controls the fidelity of
the initial global relevance estimate. We recommend allocating roughly
$10$--$20\%$ of the total frame budget to global anchors; our default
$b_0=12$ corresponds to $18.75\%$ of $B=64$. Increasing $b_0$ provides
a denser global estimate but leaves fewer frames for local selection,
while decreasing it places more of the budget under interval-level
allocation.

\subsubsection{VideoITG on \cref{tab:matchedbudget}}
VideoITG shows little change under matched budgets, with identical accuracy on VideoMME and LongVideoBench. Unlike the other selectors, VideoITG accepts at most 512 input frames, so the video is already reduced to a compact candidate set before selection, leaving less room for MetaSampling to alter the retained evidence. For a fair comparison, we impose the same 512-frame input constraint on MetaSampling rather than exposing it to additional frames, which would provide an artificial advantage.
\section{Limitations}
MetaSampling inherits the behavior of the underlying frame selector and
therefore cannot recover evidence that the selector consistently fails to
identify. Its benefit can also be smaller when the selector already operates
on a heavily reduced candidate set, as observed with VideoITG. Moreover,
MetaSampling does not guarantee an accuracy improvement for every
selector--backbone pair: while frame count is reduced in all 36 evaluated
configurations, accuracy decreases in 11 cases. Finally, our current
allocation uses fixed hyperparameters and a fixed duration threshold; adapting
these quantities to individual videos or questions remains an interesting
direction for future work.

\section{Conclusions}
We introduced MetaSampling, a training-free strategy for using existing
frame selectors more efficiently in downstream VQA. MetaSampling uses a
small global selection to allocate local sampling budgets and reuses the
same frame selector within those temporal intervals. Across 72 evaluation
runs forming 36 paired comparisons, our results suggest that improving how
existing selectors are invoked provides a complementary direction to
designing increasingly sophisticated frame-selection models.
{
    \small
    \bibliographystyle{ieeenat_fullname}
    \bibliography{main}
}


\end{document}